\documentclass[conference]{IEEEtran}
\IEEEoverridecommandlockouts                              % This command is only needed if 
\usepackage[]{algorithm2e}
\usepackage{algorithmic}
\usepackage{amsmath}
\usepackage{amssymb}
\usepackage{mathalpha}
\usepackage{caption}
\usepackage{float}
\usepackage{subfigure}
\usepackage{graphicx}
\usepackage{accents}
\usepackage[font=small]{caption}
\usepackage{array}
\usepackage{url}

\newcolumntype{M}[1]{>{\centering\arraybackslash}m{#1}}
\newcommand{\vecn}[1]{\boldsymbol{#1}}
\newcommand{\dvec}[1]{\dot{\boldsymbol{#1}}}
\newcommand{\ddvec}[1]{\ddot{\boldsymbol{#1}}}
                                                     
\title{\LARGE \bf
A Suitable Hierarchical Framework with Arbitrary Task Dimensions under Unilateral Constraints for physical Human Robot Interaction 
}

\author{Juan D. Mu\~noz Osorio$^{1,2,*}$, Felix Allmendinger$^{1}$
\thanks{ $^{1}$ Authors are with the corporate research of KUKA Germany GmbH, 86165 Augsburg, Germany}% <-this % stops a space
\thanks{ $^{2}$ Author is with the institute of mechatronic systems (IMES), Leibniz University of Hannover, 30167 Hannover, Germany}
\thanks{*This work was partly supported by the German Federal Ministry of
Education and Research (BMBF) through the project Internet of Construction
(grant no. 02P17D083.}
}

\begin{document}
\IEEEoverridecommandlockouts \IEEEpubid{\makebox[\columnwidth]{978-1-6654-8217-2/22/\$31.00~\copyright2022 IEEE \hfill} \hspace{\columnsep}\makebox[\columnwidth]{ }}
\maketitle
 \IEEEpubidadjcol
%\thispagestyle{empty}
%\pagestyle{empty}

%%%%%%%%%%%%%%%%%%%%%%%%%%%%%%%%%%%%%%%%%%%%%%%%%%%%%%%%%%%%%%%%%%%%%%%%%%%%%%%%
\setcounter{footnote}{2}

\begin{abstract}

In the last years, several hierarchical frameworks have been proposed to deal with highly-redundant robotic systems. Some of that systems are expected to perform multiple tasks and physically to interact with the environment. However, none of the proposed frameworks is able to manage multiple tasks with arbitrary task dimensions, while respecting unilateral constraints at position, velocity, acceleration and force level, and at the same time, to react intuitively to external forces. This work proposes a framework that addresses this problem. The framework is tested in simulation and on a real robot. The experiments on the redundant collaborative industrial robot (KUKA LBR iiwa) demonstrate the advantage of the framework compared to state-of-the-art approaches. The framework reacts intuitively to external forces and is able to limit joint positions, velocities, accelerations and forces.  

\end{abstract}

%%%%%%%%%%%%%%%%%%%%%%%%%%%%%%%%%%%%%%%%%%%%%%%%%%%%%%%%%%%%%%%%%%%%%%%%%%%%%%%%

\section{Introduction}

Modern robotics tend to reduce the gap between human and machines. Just like a human, a robot is expected to have high versatility and interact with the environment. The versatility of a human comes from the large number of body joints. Visualize a service task as drilling a hole. The execution of the task requires at least 5 degrees of freedom (DOF). The trajectory of the drill requires a full definition on position (3 DOF), and at least two orientation coordinates (2 DOF).\footnote{The angle around the axis of the drill is not relevant to perform the task} There exists kinematic redundancy with respect to this task. The human body can have infinity number of postures without changing the desired pose of the tool. This redundancy allows the human to simultaneously perform other tasks, e.g., optimize the posture of the body to maximize comfort. Robot controllers often solve the redundancy to achieve multiple tasks in a prioritized manner. The highest priority is given to an indispensable task (from now on, called constraints in this work). For instance, joint-limit avoidance or collision-avoidance must be accomplished to ensure a safe task execution. The lowest priority is given to tasks that are desired but are not required for performing the main purpose of an application, for instance, to use the minimum kinetic energy during motion. 

%Physical interaction between two physical systems is only ensured if one complement the other: If one system is an admittance (accept effort and yield flow), the other must be an impedance (accept flow and yield effort) and viceversa \cite{hogan1984impedance}. While the human muscles behave as impedance, the environment is admittance. Therefore. physical interaction of the robot with the environment is only ensured if the robot behaves as an impedance.  
 
A suitable control framework must be employed to command the robot to achieve multiple tasks in a hierarchical manner, while respecting all constraints. In addition, the framework must solve the redundancy considering contact between the robot structure and the human and/or environment to achieve physical human-robot interaction (pHRI). The robot must react intuitively to external forces in the task and in the null-space.  

Many frameworks has been proposed to achieve this goal, just to name a few:  \cite{Hutter.2014,AlexanderDietrich.2015,Dietrich.03.11.201307.11.2013,HierarchicalQPImpedance,Dehio.20.05.201924.05.2019}. These works focus on the response of the framework to external forces in the task space. Experiments showed that the robot follows a desired dynamic behavior in the task space. However, none of these works did a proper comparison of the behavior in the null-space when external forces disturb the robot. Although the methods may follow a desired dynamic behavior in the null-space, they may not have an intuitive reaction to a force. An intuitive response would be a motion in the direction of the applied force. Redundancy must be solved as the nature would do, following physical laws.  \cite{Bruyninckx.2428April2000} demonstrated that solvers based on null-space projectors and dynamic-consistent pseudoinverses (as \cite{Dietrich.03.11.201307.11.2013,Dehio.20.05.201924.05.2019,Kathib85,khatib2004whole}) minimize, what the authors of  \cite{Bruyninckx.2428April2000} call "acceleration energy",\footnote{Note that the concept of energy does not have the classical meaning due to the extra time derivative involved.} and follow the Gauss's  principle \cite{gauss}. Considering that Gauss's principle is equivalent to d'Alembert's principle for holonomic constraints. It can be deduced that the control force determined by minimizing the "acceleration energy" would be the force that nature would employ to solve the redundancy. "\textit{Since this principle underlies the evolution of constrained motion in mechanical systems in nature}" \cite{Udwadia.2003}. The drawback of these solvers is that they do not allow a direct inclusion of inequalities. Although these solvers can be modified to include inequalities at position, velocity and acceleration levels (as shown in \cite{SJSus,SCS}), inclusion of torque limits has not been treated. Contrary to those solvers, quadratic programming (QP)-based solvers as \cite{HierarchicalQPImpedance,del2015prioritized,DelPrete.2018,JuanQuiroz} allow inclusion of torque limits. However, the response to external forces in task and null-space has not been properly evaluated. We are aware of many other multitasks frameworks that allow arbitrary dimension of task and inequality constraints as the iTasC \cite{Smits.2008,Decre.2009,Vanthienen.2012}. Nevertheless, these frameworks result in a low-level velocity controller. A stable physical interaction requires the robot to behave as an impedance \cite{hogan1984impedance}, which requires joint force/torque control.    

The main contribution of this work is the development of a hierarchical framework that combines the advantages of null-space projector and QP-based redundancy solvers. The framework: 
\begin{itemize}
\item solves the redundancy as nature would do producing an intuitive behavior of the robot,
\item allows limitation of the torques,
\item properly considers external forces/torques for limitation of position and velocities, reducing the risk of violating the limits due to physical interaction of the robot with the environment. 
\end{itemize} 

\section{Problem definition}

Consider a manipulator with $n$ number of joints that must perform  k-number of tasks under l-number of unilateral constraints.  Each task can be defined by desired accelerations $\ddvec{x}_{\textrm{d},i} \in \mathbb{R}^{m_i}$, where the sub-index []$_i$ indicates the order of priority from the most important 1 to the less important k. The desired acceleration of each task is commonly defined given the error dynamics $\ddvec{x}_{\textrm{d},i} + \vecn{K}_i (\Delta \boldsymbol{x}_i) +\vecn{D}_i(\Delta \dvec{x}_i)$, where $\Delta \boldsymbol{x}_i$ is the error between the desired and the current position $\boldsymbol{x}_{\textrm{d},i} - \vecn{x}_i$. $\vecn{K}_i$ and $\vecn{D}_i$ are the positive-definite stiffness and damping matrices of each task, respectively. If $\vecn{q} \in \mathbb{R}^{n}$ is the corresponding joint configuration vector, there is a forward kinematics function that relates $\vecn{q}$ to the position in the task space, such that $\textrm{FK}(\vecn{q}) = \vecn{x}_i$. Assume the redundant case where the manipulator is redundant with respect to at least the main task ($n > m_1$), where $m$ is the dimension of the task.

The redundancy problem is commonly solved by two different approaches: using null-space projectors and pseudoinverses, and writing the problem as a quadratic cost function to be solved by quadratic programming solvers. 

On the one side, \cite{Peters.2008} shows that the use of the dynamic consistent pseudoinverse of the Jacobian (as proposed in the Operational Space Control (OSC) \cite{khatib1980commande}) is the solution of minimizing the control input using the inverse of the inertia matrix as  metric $\vecn{M}^{-1}$, under the constraint of having zero error in the task t.
\begin{subequations}\label{eq:OSCminProblem}
\begin{alignat}{2}
&\!\min_{\boldsymbol{\tau}}        &\qquad&  \frac{1}{2}\vecn{\tau}^T \vecn{M}^{-1} \vecn{\tau} \label{eq:OSCmin}\\
&\text{subject to} &      & \vecn{J}_\textrm{t} \vecn{M}^{-1} \vecn{\tau} =  \ddvec{x}_\textrm{d,t}-\dvec{J}_\textrm{t}\dvec{q} - \vecn{J}_\textrm{t} \vecn{M}^{-1}\vecn{\nu},\label{eq:OSCminC}
\end{alignat}
\end{subequations}
 where $\vecn{\tau}$ is the vector of generalized joint forces to command the robot. The Jacobian $\vecn{J}_\textrm{t}$ maps the velocities from the task space to the joint space. $\vecn{\nu}$ is the vector of coriolis forces. The gravity vector $\vecn{g}$ is added to the command vector to compensate gravity after computing the optimal solution to the problem.

%And if the external forces are not ignored. The complete minimization problem becomes
%\begin{subequations}\label{eq:OSCminProblem}
%\begin{alignat}{2}
%&\!\min_{\boldsymbol{\tau}}        &\qquad&  \vecn{\tau}^T \vecn{M}^{-1} \vecn{\tau} \label{eq:OSCminD}\\
%&\text{subject to} &      & \vecn{J} \vecn{M}^{-1} \vecn{\tau} =  \ddvec{x}_\textrm{t}-\dvec{J}\dvec{q} - \vecn{J} \vecn{M}^{-1}\vecn{\nu} - \vecn{J} \vecn{M}^{-1} \vecn{J}^T \vecn{f} _\textrm{ext}\label{eq:OSCminE}
%\end{alignat}
%\end{subequations}
As stated in the introduction, this solution follows the Gauss principle \cite{gauss} and minimizes the "acceleration energy" given by:
\begin{equation}
E_\textrm{acc} = \frac{1}{2} \vecn{\tau} ^T \vecn{M}^{-1} \vecn{\tau}.
\end{equation}
Minimizing the "acceleration energy" follows the principle of constrained motion in nature to solve the redundancy. Thus, an approach has a more intuitive reaction when spending less "acceleration energy" during motion. The energy minimization can be, therefore, used as a metric to determine an intuitive reaction.
 
On the other site, QP-based solvers minimize the error between the current and the desired acceleration in the task space. The optimization problem is written as: 
\begin{subequations}\label{eq:optProbGes}
\begin{alignat}{2}
&\!\min_{\ddvec{q},\boldsymbol{\tau}}        &\qquad&  \| \boldsymbol{J}_\textrm{t}\ddvec{q} - (\ddvec{x}_\textrm{d,t}-\dvec{J}_\textrm{t}\dvec{q}) \| ^{2} + wr\label{eq:optProb}\\
&\text{subject to} &      & \boldsymbol{\tau} =  \boldsymbol{M}\ddvec{q} + \vecn{\nu} + \vecn{g}.\label{eq:constraint1}
\end{alignat}
\end{subequations}
The constraint \ref{eq:constraint1} keeps the dynamic consistency between the generalized forces $\boldsymbol{\tau} $ and the joint accelerations $\ddvec{q}$. The weight $w$ weighs a regularization term $r$ that regularizes the accelerations in the null-space to have a unique solution. 
The regularization term $r$ commonly minimizes: 
\begin{itemize}
\item the norm of the acceleration error for performing a joint position task, i.e., $r = \| \ddvec{q} - \ddvec{q}_\textrm{d} \| ^{2}$, where $\ddvec{q}_\textrm{d} = \vecn{K}(\vecn{q}_\textrm{d} - \vecn{q}) - \vecn{D}\dvec{q}$ \cite{del2015prioritized} or
\item the norm of the torques, i.e., $r =\| \vecn{\tau} \|$ \cite{HierarchicalQPImpedance} or
\item the norm of the joint accelerations, i.e., $r =\| \ddvec{q} - \ddvec{q}_\textrm{d} \| ^{2}$, where $\ddvec{q}_\textrm{d} = 0$ \cite{JuanQuiroz}.
\end{itemize}

Having a desired joint position means that a desired robot's posture is enforced in the null-space. This choice brings two problems: changing the posture by means of external forces in the null-space of the tasks is not possible,\footnote{The DOF not used to solve the main task  do not behave as passive unactuated DOF.} and a proper posture must be found depending on the application.  The other two commonly used regularization terms are not meant to reduce the kinetic or the "acceleration energy". A better approach to define the regularization term is having a damping task in the joint space, which dissipates the kinetic energy: $r =\| \ddvec{q} - \ddvec{q}_\textrm{d} \| ^{2}$, where $\ddvec{q}_\textrm{d} = -\vecn{D}\dvec{q}$. However, the motion of the robot may not be intuitive by application of external forces. A combination of the projector- and QP-based approaches takes the advantages of both. %The solver presented in \cite{Liu.2016} is a first mix of both approaches, but it stills does not minimize the acceleration energy.  

\section{The Dynamically-consistent Constrained Task Hierarchy Solver (DCTS)}\label{sec:NewApproach}

It might appear simple to combine the two methods by solving the problem in eq. \ref{eq:OSCminProblem} with a QP solver. That approach would avoid numerical singularities, and it opens the possibility to directly include unilateral constraints in the solver. However, three issues would remain:
\begin{itemize}
\item  inclusion of new constraints is only possible at the same level of priority than the task described by $\ddvec{x}_\textrm{d,t}$ in eq. \ref{eq:OSCminC},
\item external forces are not considered and 
\item inclusion of multiple tasks with different levels of priority is not possible.
\end{itemize}

\subsection{Separation of the task from the constraints}

\cite{Flacco.5620135102013} treats the solution of redundancy at velocity level under inequalities for multiple tasks. The approach is formatted as a QP problem, where the tasks get different priority level by adding a task-scaling factor $s$. This factor scales the task in case joint limits hinder a fully accomplishment of the task. In this way, the direction of the error vector is maintained and the problem becomes solvable. The scaling factor is then maximized (i.e., equal to one if the task is feasible and less than one if the saturated joints are needed to fully perform the task). An adaptation of this approach at dynamic level solves the first issue previously mentioned. The quadratic problem in eq. \ref{eq:OSCminProblem} becomes 
\begin{subequations}\label{eq:OSCminProblemS}
\begin{alignat}{2}
&\!\min_{\ddvec{q},\boldsymbol{\tau},s}        &\qquad&  \frac{1}{2}\vecn{\tau}^T \vecn{M}^{-1} \vecn{\tau} + w(1-s)^2\label{eq:OSCminDS}\\
&\text{subject to} &      & \boldsymbol{\tau} =  \vecn{M} \ddvec{q}+ \vecn{\nu} + \vecn{g}\label{eq:const1}\\
&                  &      & \boldsymbol{\tau}_\textrm{min} \leq  \boldsymbol{\tau} \leq \boldsymbol{\tau}_\textrm{max} \label{eq:const2}\\
&                  &      & \ddvec{c}_\textrm{min}  \leq  \vecn{J}_\textrm{c} \ddvec{q} + \dvec{J}_\textrm{c} \dvec{q} \leq \ddvec{c}_\textrm{max}\label{eq:const3}\\
&                  &      & \vecn{J}_\textrm{t} \ddvec{q} =  s\ddvec{x}_\textrm{d,t}-\dvec{J}_\textrm{t}\dvec{q}, \label{eq:OSCminES}
\end{alignat}
\end{subequations}
where $w \gg 1$ is a large penalty factor to give a scaling factor maximization a higher priority than the minimization of the "acceleration energy".% Note that at velocity level the scaling factor can never become less than 0, because a zero desired task velocity means no motion in the operational space. At acceleration level, however, the scaling factor can become less than 0, because zero acceleration in the operational space means constant velocity, i.e. if the velocity needs to be reduced the acceleration should change is direction with a negative scaling factor.  

Constraint \ref{eq:const3} is used to include linear unilateral constraints defined in a "limited space" (space in which the inequalities are included). $\boldsymbol{J_\textrm{c}} \in \mathbb{R}^{l \times n}$ relates the velocity on the limited space with the joint velocities ($\dvec{c} = \vecn{J}_\textrm{c} \dvec{q}$), $\boldsymbol{c} \in \mathbb{R}^{l} = \textrm{FK}(\vecn{q})$, $\dvec{c}$ and $\ddvec{c}$ are the position, velocity and acceleration of the limited directions. $\textrm{FK}(\vecn{q})$ is a forward kinematics function that computes the position of a point of interest in the limited space. If the limited space is the joint space, the inequality \ref{eq:const3} becomes $\ddvec{q}_\textrm{min} \leq  \ddvec{q} \leq \ddvec{q}_\textrm{max}$. The optimal solution guarantees $\boldsymbol{\tau}$ and $\ddvec{c}$  to be between their maximal and minimal limits in constraint \ref{eq:const2} and \ref{eq:const3} respectively (where $\ddvec{c} = \vecn{J}_\textrm{c} \ddvec{q} + \dvec{J}_\textrm{c} \dvec{q}$). $\ddvec{c}_\textrm{max}$ and $\ddvec{c}_\textrm{min}$ are shaped to include velocity and position limits, as proposed in \cite{SJSus,DelPrete.2018,SCS}.  
 
\subsection{Consideration of external forces}

\cite{AlexanderDietrich.2015} pointed out that controllers based on the widely used  operational space control \cite{khatib1980commande} commonly ignore the external forces in the task dynamic behavior. Neglecting these forces implies to have a different dynamic behavior than the desired one. \cite{AlbuSchaffer.1419Sept.2003} shows that the implementation of an impedance behavior in the task without shaping the desired inertia avoids the inclusion of these forces. This implementation can be included in the framework proposed here by computing the desired acceleration as $\ddvec{x}_\textrm{d} = \vecn{JM}^{-1}\vecn{J}^\textrm{T}\vecn{f}_\textrm{d}$, where  $\vecn{f}_\textrm{d} = \vecn{K} (\Delta \boldsymbol{x}_i) +\vecn{D}(\Delta \dvec{x}_i)$.\footnote{The damping should be computed as a function of the desired inertia matrix as proposed in \cite{AlbuSchaffer.1419Sept.2003}} However, the framework is not restricted to it. If a desired inertia shaping must be included, external forces can be included in the problem. Constraint \ref{eq:OSCminES} becomes
\begin{equation}
\vecn{J}_\textrm{t} \ddvec{q} =  s\ddvec{x}_\textrm{d,t}-\dvec{J_\textrm{t}}\dvec{q} -\vecn{J}_\textrm{t} \vecn{M}^{-1}\vecn{\tau}_\textrm{ext} 
\end{equation}
$\vecn{\tau}_\textrm{ext}$ is the vector of external generalized joint forces that can be measured by joint force/torque sensors.\footnote{Some industrial collaborative robots include these sensors as: the KUKA LBR iiwa \cite{LBR}, the Yuanda \cite{YUANDA}, the Panda emika \cite{PANDA}, the sawyer \cite{SAWYER} and the IRB 14000 YuMi \cite{YUMI}} Using this approach, the measurement of the external force in the task space is avoided. Note that this is a different approach than the proposed in \cite{Liu.2016}, where the external forces are directly included in constraint \ref{eq:const1}. That approach does not allow physical interaction. The cancellation of external forces by adding them to the command vector produces zero acceleration, i.e. no motion, if the robot is in a static state. 

Additionally, external forces must be also considered in the unilateral constraints. Otherwise, a violation of the limits can occur during physical interaction. This is the main drawback of the approach proposed in \cite{DelPrete.2018}. The unilateral constraint \ref{eq:const3} becomes
\begin{equation}
\ddvec{c}_\textrm{min} - \vecn{J}_\textrm{c}\vecn{M}^{-1}\vecn{\tau}_\textrm{ext} \leq  \vecn{J}_\textrm{c} \ddvec{q} + \dvec{J}_\textrm{c} \dvec{q} \leq \ddvec{c}_\textrm{max}- \vecn{J}_\textrm{c}\vecn{M}^{-1}\vecn{\tau}_{\textrm{ext}}
\end{equation}

\subsection{Extension to multiple tasks}

Having multiple tasks increases the complexity of the problem, because every task should be solved reducing the "acceleration energy". At the same time, tasks must be performed in a hierarchical way. Based on the mixed solver proposed in \cite{Liu.2016}, a hierarchy scheme can be achieved while minimizing the "acceleration energy". The inclusion of dynamically consistent null-space projectors ($\vecn{N}_i = \vecn{I} - \vecn{J}^{\textrm{T}}_i\vecn{\bar{J}}^{\textrm{T}}_i$) guarantees also that  kinetic energy is not inserted in the null-space. That consequence is explained by the use of the  dynamically consistent pseudo inverse $\vecn{\bar{J}} = \vecn{M}^{-1}\vecn{J}(\vecn{JM}^{-1}\vecn{J}^\textrm{T})^{-1}$. This inverse is "\textit{the unique operator that solves the redundancy without injecting energy in the null-space components of motion}" \cite{Bruyninckx.2428April2000}. To include the null-space projectors, the minimization problem becomes:
\begin{subequations}
\begin{alignat}{2}
&\!\min_{\ddvec{q}^\textrm{aug},\boldsymbol{\tau},\boldsymbol{s}}        &\qquad&  \frac{1}{2}\vecn{\tau}^T \vecn{M}^{-1} \vecn{\tau} + \sum_{i=1}^{k}w_i(1-s_i)^2
 \label{eq:optProbMix}\\
&\text{subject to} &      & \boldsymbol{\tau} =  \vecn{M}\ddvec{q}+ \vecn{\nu} + \vecn{g}\label{eq:constraintMix}\\
&                  &      & \boldsymbol{\tau}_\textrm{min} \leq  \boldsymbol{\tau} \leq \boldsymbol{\tau}_\textrm{max} \label{eq:const4}\\
&                  &      & \ddvec{c}_\textrm{min} - \vecn{J}_\textrm{c}\vecn{M}^{-1}\vecn{\tau}_\textrm{ext} \leq \vecn{J}_\textrm{c} \ddvec{q} + \dvec{J}_\textrm{c} \dvec{q} \leq \notag\\ 
&				   &     & \ddvec{c}_\textrm{max}- \vecn{J}_\textrm{c}\vecn{M}^{-1}\vecn{\tau}_{\textrm{ext}} \label{eq:const5}\\
&                  &      & \boldsymbol{J}^\textrm{aug} \ddvec{q}^\textrm{aug} =  \vecn{S}^\textrm{aug}\ddvec{x}^\textrm{aug} -\dvec{J}^\textrm{aug}\dvec{q} - \notag\\
&				   &      &\vecn{J}^\textrm{aug} \vecn{M}^{-1}\vecn{\tau} _\textrm{ext}\\
%&                  &      & \ddot{q}_{min} \leq  \ddot{q} \leq \ddot{q}_{max} \label{eq:constraint2} \\
&                  &      & \ddvec{q} = \vecn{N}  \ddvec{q}^\textrm{aug} = \sum_{i=1}^{k}\vecn{N}_\textrm{i} \ddvec{q}_i  \label{eq:constraintMix}
\end{alignat}
\end{subequations}
with $\ddvec{q}^\textrm{aug}  = \begin{bmatrix}
\ddvec{q}_1\\
\vdots \\
\ddvec{q}_k
\end{bmatrix}$, $\ddvec{x}^\textrm{aug}  = \begin{bmatrix}
\ddvec{x}_1\\
\vdots \\
\ddvec{x}_k
\end{bmatrix}$, $\vecn{J}^\textrm{aug}  = \begin{bmatrix}
\boldsymbol{J}_\textrm{1}\\
\vdots \\
\boldsymbol{J}_{k}
\end{bmatrix}$, $\boldsymbol{N} = [\vecn{N}_1 \hdots \vecn{N}_k]$. The diagonal matrix $\vecn{S}^\textrm{aug}$ contains diagonal values from $s_1...s_k$. To scale every task in a prioritized order  the weights are set $w_i \gg w_{i-1}$. The strict hierarchy between tasks is guaranteed by using augmented null-space projectors. When a limit is encountered and is in a conflict with more than one task, tasks may be not scaled in a strict-priority order. The choice of the weights gives this strictness.

The use of a QP solver brings the advantage of avoiding singularities when the Jacobian losses rank due to the avoidance of inverses computation. However, in this framework, the computation of $\vecn{\bar{J}}$ is required. To avoid singularities many of the methods in the literature may be implemented \cite{khatib2004whole,Deo1995,HanSingAndJoint}. The best choice to keep the use of QP solvers and to avoid singularities is to formulate the computation of  $\vecn{\Lambda} = (\vecn{JM}^{-1}\vecn{J}^\textrm{T})^{-1}$ as a least-square problem, as proposed in \cite{HierarchicalQPImpedance}. 

\section{Simulation and experiments}

This section compares state-of-the-art methods with the DCTS and demonstrates the problem of solving a QP problem that does not minimize the "acceleration energy". Simulations and experiments demonstrate the advantages of the proposed framework. The tests are performed on a KUKA LBR iiwa ($n = 7$). The values of the limits used in simulations and experiments are shown in table \ref{tab:limFeas}. 
   
\begin{table}
\begin{center}
\begin{tabular}{ |c||c|c|c|c|c|c|c|} 
 \hline
\makebox[3mm]{Joint}
&\makebox[3mm]{$q_1$}&\makebox[3mm]{$q_2$}&\makebox[3mm]{$q_3$}
&\makebox[3mm]{$q_4$}&\makebox[3mm]{$q_5$}&\makebox[3mm]{$q_6$}&\makebox[3mm]{$q_7$}\\\hline\hline
$\vecn{q}_\textrm{max}$ [deg] &165&115&165&115&165&115&170\\\hline
$\vecn{q}_\textrm{min}$ [deg]  &-165&-115&-165&-115&-165&-115&-170\\\hline
$\vecn{v}_\textrm{max}$ [deg/s]&100&110&100&130&130&180&180\\\hline
$\vecn{v}_\textrm{min}$ [deg/s]&-100&-110&-100&-130&-130&-180&-180\\\hline
$\vecn{\tau}_\textrm{max}$ [Nm]&100&100&100&100&100&100&100\\\hline
$\vecn{\tau}_\textrm{min}$ [Nm]&-100&-100&-100&-100&-100&-100&-100\\\hline
\end{tabular}
\end{center}
\caption{Robot joint limits used in the experiments}
\label{tab:limFeas}
\end{table}
%This issue becomes more critical, when the task is desired only in rotational coordinates and the translation is free to be resolved by the redundancy algorithm.

\subsection{Simulation results}

The simulations were run in Matlab/Simulink using the qpOASES library \cite{ferreau2014qpoases} as a QP solver. The first and second simulations compare the OSC, the QP-MT and QP-MT. The purpose of these simulations is to show the advantage of minimizing the "acceleration energy" when task redundancy exists. Thus, the task for the end effector center is defined in only two rotational coordinates ($m = 2$), i.e., the null-space of the Cartesian space has 4 DOF (1 rotational and 3 rotational coordinates). In the first simulation, the end effector is expected to rotate 15 degrees from the initial pose around the x-axis of the world frame, while keeping the orientation around the y-axis. The redundancy is solved by three approaches:
\begin{itemize}
\item the OSC, i.e., solving the QP problem in eq. \ref{eq:OSCminProblem}
\item solving the problem in eq. \ref{eq:optProbGes}, with a regularization term that minimizes the norm of the torque  $r =\| \vecn{\tau} \|$ (QP-MT)
\item solving the problem in eq. \ref{eq:optProbGes}, with a regularization term that minimizes the kinetic energy in joint space $r =\| \ddvec{q} - \ddvec{q}_\textrm{d} \| ^{2}$, where $\ddvec{q}_\textrm{d} = -\vecn{D}\dvec{q}$ (QP-MD) 
\end{itemize} 

\begin{figure}
\centering
        \begin{center}
	    \subfigure[]{\includegraphics[scale=0.52]{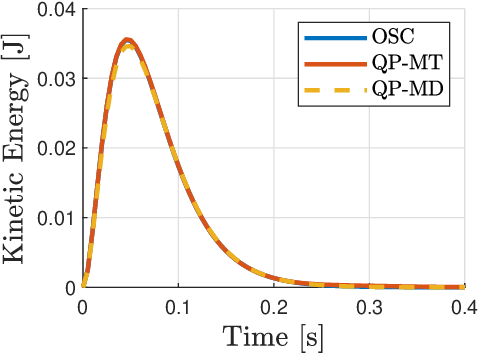}\label{fig:kinEneNoForce}} \subfigure[]{\includegraphics[scale=0.52]{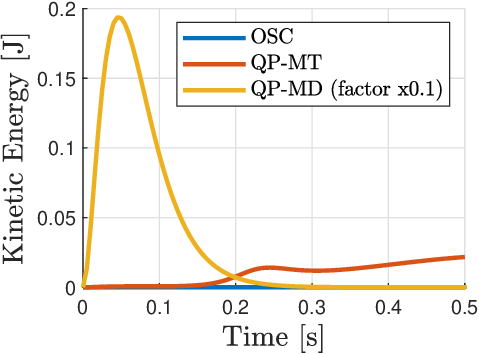}\label{fig:kinEnergyNullSpace}}
        \caption[]{Kinetic energy used to solve a 2D rotational task. a) Energy in task space. b) Energy in null-space. 
        }\label{fig:EnergysPlotPHRITestNoExtForce}
        \end{center}
\end{figure}

All three approaches solve the task and minimize the error to zero in the desired coordinates (the plot is not shown for sake of space). Fig. \ref{fig:kinEneNoForce} shows that the kinetic energy in task space is approximately the same for all the approaches and converges to zero. The kinetic energy in the null-space, though, looks completely different for all the three solvers (see fig.\ref{fig:kinEnergyNullSpace}). While the OSC shows 0 energy, QP-MT shows a energy that increases over time. This increment of the energy is because minimizing the torque does not guarantee that the velocity becomes zero, i.e., energy is not dissipated. If a damping task is used (QP-MD), the kinetic energy in the null-space is reduced but the total energy spent is much higher than minimizing the norm of the torque. These results demonstrate that the dynamically consistent pseudo-inverse does not inject energy in the null-space components of motion. As QP-MT and QP-MD solvers take a solution that minimizes the error in the task space without using this inverse, kinetic energy may be injected and it may not be dissipated. The dissipation depends on the chosen regularization term. 
	
The second simulation illustrates the different responses of the approaches to external forces. Envision a worker who must teach the robot different points located on a table, where the robot must automatically drill after. Moving the robot with a visual user interface may be slower and less intuitive than moving the robot hand-guided. To reduce the complexity of the teaching process, the end effector must keep a desired orientation around two axes to have the drill perpendicular to the table. The position and the orientation around the drill axis are not controlled. Application of external forces should move the robot along these coordinates. This application is a use-case of the task previously simulated. The end effector starts with an orientation that positions the drill perpendicular to the table. An external force of 10 N is applied on the end effector in +x direction for 400 ms. The force disturbs the end effector making it lose its desired orientation. The motion must follow the desired dynamic-behavior in task space.   
\begin{figure}
\centering
        \begin{center}
	    \subfigure[]{\includegraphics[scale=0.52]{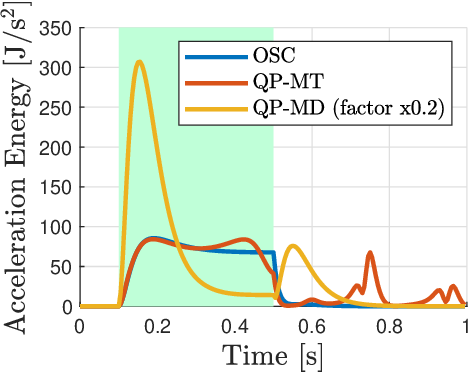}\label{fig:AccEnergy}} \subfigure[]{\includegraphics[scale=0.52]{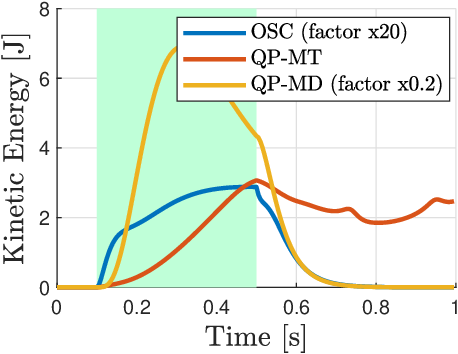}\label{fig:KinEnergy}}
	
        \caption[]{Comparison of minimization functions to solve redundancy. a) Acceleration energy. b) Kinetic Energy. The green zone represents the time interval when the external force was applied.
        }\label{fig:EnergysPlotPHRITest}
        \end{center}
\end{figure}
\begin{figure}[H]
\centering
        \begin{center}
	    \subfigure[]{\includegraphics[scale=0.52]{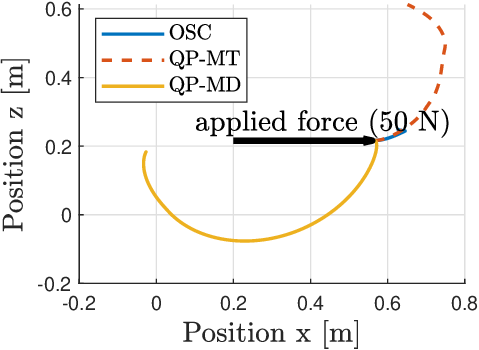}\label{fig:motion}} \subfigure[]{\includegraphics[scale=0.52]{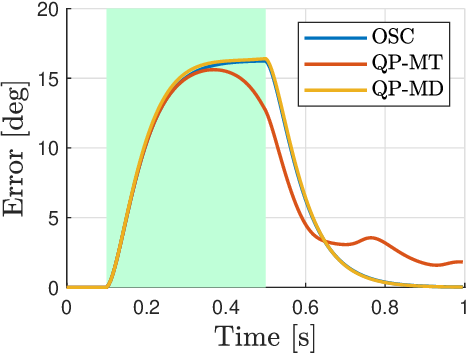}\label{fig:error}}
	
        \caption[]{Comparison of minimization functions to solve redundancy. a) Resulting motion in Cartesian null-space. b) Task error. The green zone represents the time interval when the external force was applied.
        }\label{fig:MotionAndError}
        \end{center}
\end{figure}
As expected, the OSC which minimizes directly the "acceleration energy" behaves more intuitively. Figure  \ref{fig:EnergysPlotPHRITest} shows that the external force injected more kinetic energy in the robot using QP-MT and QP-MD, despite of having the same task as OSC. Although OSC and QP-MD reduced the error in the task space having a similar behavior  (see fig. \ref{fig:error}), only the OSC had an intuitive motion (see fig. \ref{fig:motion}). As the human expects, by pushing a unit rigid body that is in an initial still state, the acceleration should produce a motion in the direction of the force times the inertia matrix $\boldsymbol{\Lambda}$.\footnote{the inertia matrix may have coupling terms that deviates the direction of motion from the direction of the force} This intuitive response only occurs using OSC, the end effector moves following the Newton's law $\ddvec{x} = \boldsymbol{\Lambda} ^{-1} \boldsymbol{f}_\textrm{ext}$ in the null-space (translation). %In the task space, the motion follows the motion equation in \ref{eq:dynSys}. 

\begin{figure}
\centering
        \begin{center}
        \includegraphics[scale=0.57]{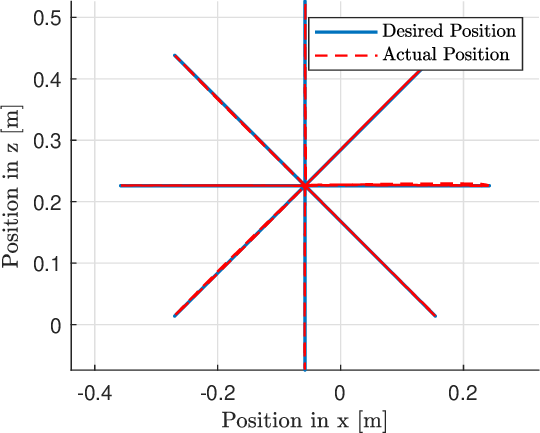}\caption[]{End effector Cartesian position using the projector-based approach} \label{fig:starPathOSC}
        \end{center}
\end{figure}

\begin{figure}
\centering
        \begin{center}
        \includegraphics[scale=0.57]{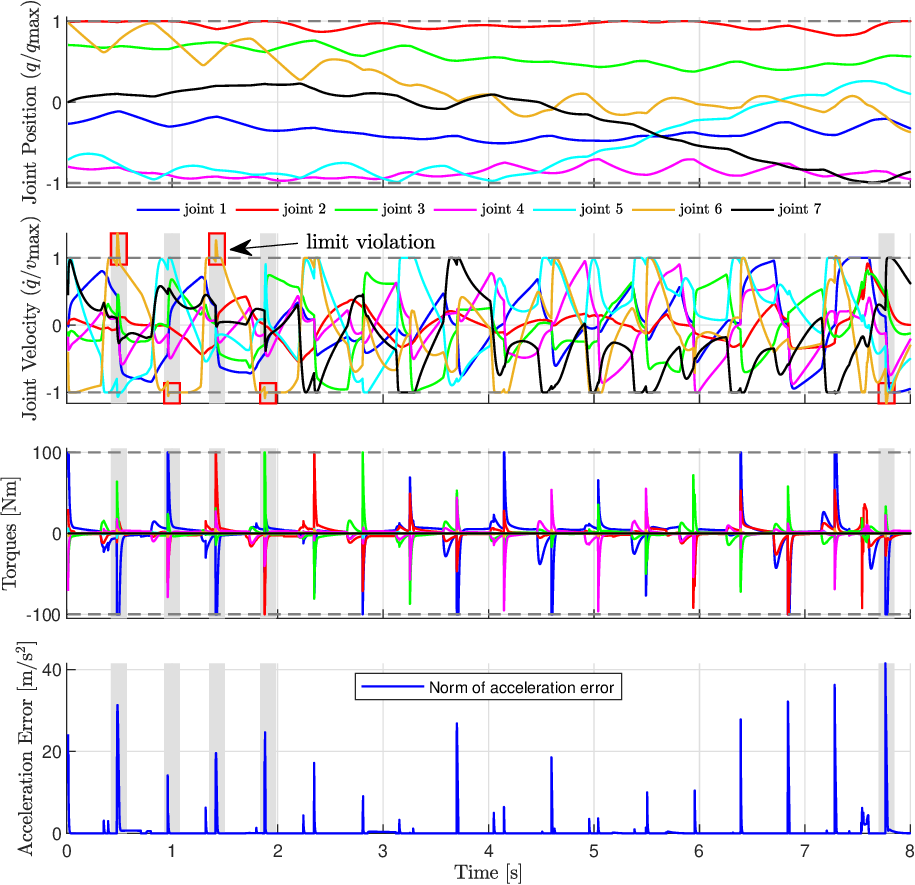}\caption[]{Results of drawing a star with the end effector using the projector-based approach. The red boxes show the velocity limits violation and the gray zones show the time-intervals in which the torque saturation is in conflict with the velocity limitation} \label{fig:OSCStartResults}
        \end{center}
\end{figure}
The next simulation demonstrates the advantages of the DCTS approach over the projector-based approaches based on OSC. The center point of the end effector should go through a series of Cartesian points connected by linear paths, starting from $\vecn{q}_\textrm{ini} =  [-0.78,2.05,2.07,-1.65,-2.08,2.03,0]^{\textrm{T}}$ [rad]. The points are vertices of an octagon inscribed in a circle lying in the (y,z) vertical plane of radius 0.3 [m]. The center of the circle lies in [0.55,-0.05,0.22] [m]. The end effector starts in the center of the circle and goes to every vertice of the octagon, and returns to the center every time. 
\begin{figure}
\centering
        \begin{center}
        \includegraphics[scale=0.57]{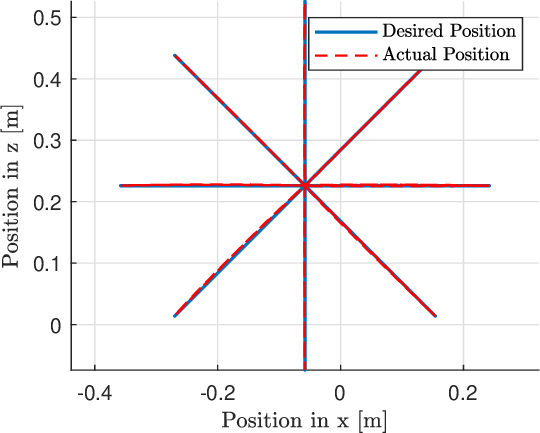}\caption[]{End effector Cartesian position using DCTS} \label{fig:starPathNew}
        \end{center}
\end{figure}
\begin{figure}
\centering
        \begin{center}
        \includegraphics[scale=0.57]{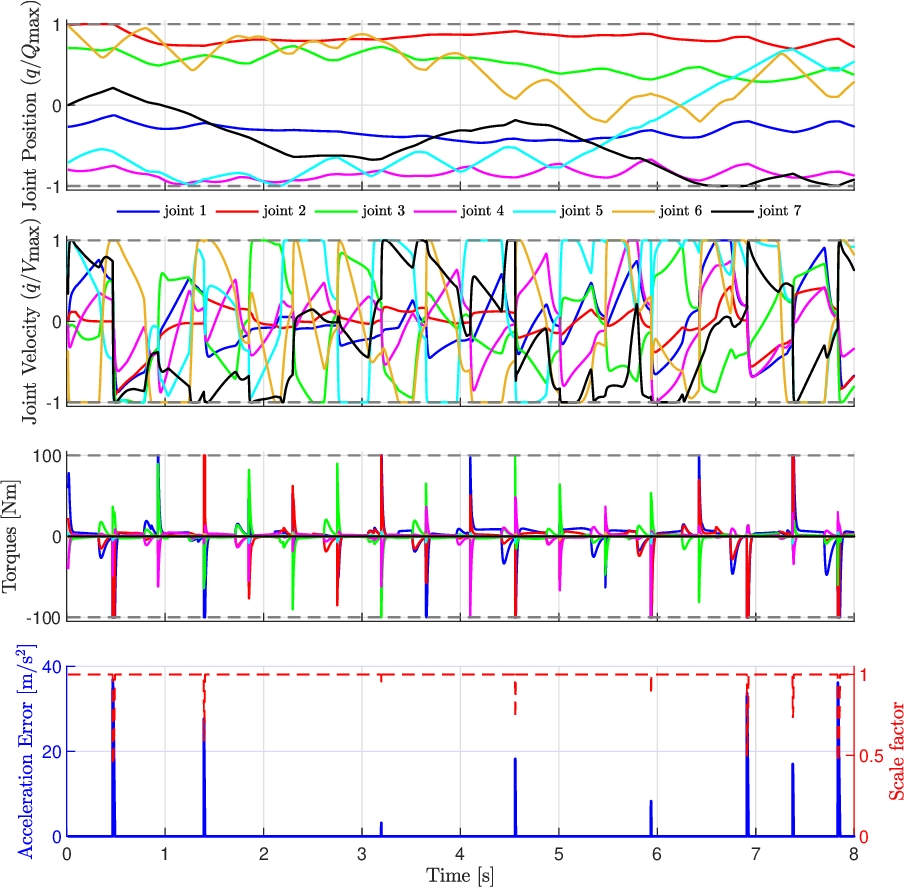}\caption[]{Results of drawing a star with the end effector using DCTS} \label{fig:NewStartResults}
        \end{center}
\end{figure}
A primary task of dimension $m_1 = 3$ is specified (only Cartesian positions). The desired acceleration $\ddvec{x}_1$ is chosen so as to head toward the next Cartesian point with a speed 0.85 [m/s], following the desired dynamic behavior 
\begin{equation}
\ddvec{x}_\textrm{1} = -k_\textrm{v}(\dvec{x}_{c}-v\dvec{x}_\textrm{d}),
\label{Position1}
\end{equation}
where $\dvec{x}_\textrm{d} = k_\textrm{p}k^{-1}_\textrm{v}\Delta\boldsymbol{x}$ is the desired velocity. $v$ is calculated as follows:
\begin{equation}
v = \textrm{min}(1,\frac{v_\textrm{sat}}{\left \| \dvec{x}_\textrm{d} \right \|^2}).
\label{Position3}
\end{equation}
The position and velocity gains $k_\textrm{p}$ and $k_\textrm{v}$ are set to 400 s$^{-2}$ and 40 s$^{-1}$ respectively. $v_\textrm{sat}$ is 0.4 m/s.

This trajectory is particularly demanding at the vertices (reached with a tolerance of 0.001 [m]) of the octagon, where large accelerations are required to suddenly change direction. A limitation of the joint positions and velocities is included following the method in \cite{SJSus}. For the projector based approach, a naive\footnote{Naive in this context means that the torques are saturated without computing a new solution to have an optimal result. Therefore, the desired accelerations in the operational space or in the joint space will not be produced} saturation of the torque is used. Each computed command torque value $\tau_j$ is saturated to its maximal or minimal value and sent to the robot. If $\tau_{j} > \tau_{\textrm{max},j}$, then $\tau_{j} = \tau_{\textrm{max},j}$. If $\tau_{j} < \tau_{\textrm{max},j}$, then $\tau_{j} = \tau_{\textrm{min},j}$ with $j = 1...n$. 

Figures \ref{fig:starPathOSC} and \ref{fig:OSCStartResults} show the result of task execution using the projector-based approach of \cite{SCS}. Figure \ref{fig:starPathOSC} shows that the desired path in position is followed despite the limitation of joint positions, velocities and torques. However, there are two aspects to remark, given by the naive saturation of the torques:
\begin{itemize}
\item when the torque to limit the joint velocities is bigger than the maximal allowed torque, the velocity limit is not respected. The time-intervals represented by the gray zones in Fig.\ref{fig:OSCStartResults} show these cases. 
\item when the torques to produce the desired acceleration in the operational space are bigger than the maximal torques, the acceleration error increases. See Fig. \ref{fig:OSCStartResults}.
\end{itemize}   
 
The simulation was repeated using the novel proposed framework in section \ref{sec:NewApproach} (DCTS). Fig. \ref{fig:starPathNew} shows the results of the position tracking of the star. The end effector followed the desired path despite all limitations. The error of the position tracking compared to the projector-based approach is slightly less (see Tab. \ref{tab:errors}). However, the results in Fig. \ref{fig:NewStartResults} show that all limitations were respected. The error of the acceleration only increased when the task had to be scaled (scaling factor smaller than one). Such case came, when the torque limits were in conflict with the task, i.e., there was no solution to fully accomplish the task and to keep the torques between their limits. In average, the acceleration error decreased compared to the projector-based approach (see Tab. \ref{tab:errors}).  
    
%As the QP based approach solves the redundancy by minimizing the acceleration error in the Cartesian space, it can not guarantee that kinetic energy is not injected in the null space components of motion. The method that propose a damping task as regularization term dissipates the kinetic energy as fast as the OSC, with the difference that the kinetic energy injected was more than 200 times bigger leading to a non-intuitive motion. Minimizing the norm of the torque shows a similiar acceleration energy behavior during the injection of a external force. However, the kinetic energy did not get reduced as fast as with the other two methods. There is also 20 times more kinetic energy injected than with the OSC. 

 %This issue is not noticeable when the task is defined in all the available Cartesian coordinates. However, as the aim of this thesis is to have a general framework that allows to define the task in any coordinates, the "acceleration energy" must be minimized. 

%\begin{figure}[hbt!]
%\centering
%        \begin{center}
%	    \includegraphics[scale=0.58]{Images/testpHRI/%AccEnergyNoForce.eps}\label{fig:}	
%        \caption[]{Acceleration energy used to solve a 2D %rotational task. 
%        }\label{fig:EnergysPlotPHRITestNoExtForce}
%        \end{center}
%\end{figure}

  \begin{figure}
\centering
        \begin{center}
        \includegraphics[scale=0.57]{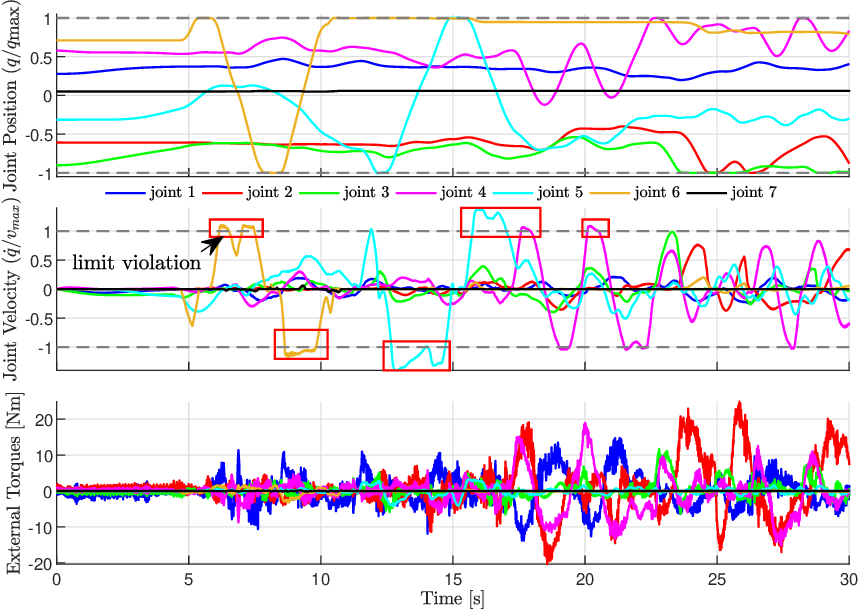}\caption[]{Results of not considering the external forces when the human moves the robot by hand} \label{fig:NoForceIncluded}
        \end{center}
\end{figure}

 \begin{table}
\begin{center}
\begin{tabular}{ |c||c|c|} 
 \hline
\makebox[10mm]{Approach}
&\makebox[18mm]{Position Error}&\makebox[18mm]{Acceleration Error}\\\hline\hline
\cite{SCS} & 0.018 [m] & 0.0294 [m/s$^2$]\\\hline  
DCTS   & 0.016 [m]	&  0.0248 [m/s$^2$]\\\hline
\end{tabular}
\end{center}
\caption{Averages errors during tracking of a star path}
\label{tab:errors}
\end{table}
\subsection{Experimental results}

Two experiments are performed on the real robot. The first experiment shows the advantage of including the external forces in the unilateral constraints of the solver. The human starts to move the robot by hand. The human intention is to drive the robot joints to their maximal and minimal limits by moving the robot as fast as possible.

\begin{figure}
\centering
        \begin{center}
        \includegraphics[scale=0.57]{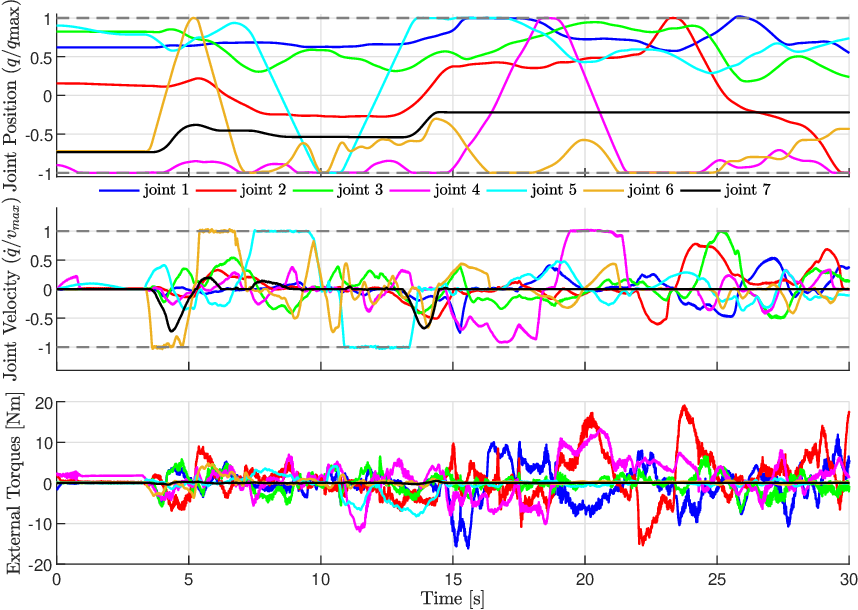}\caption[]{Results of considering the external forces when the human moves the robot by hand} \label{fig:ForceIncluded}
        \end{center}
\end{figure}

Figures \ref{fig:NoForceIncluded} and \ref{fig:ForceIncluded} show the results of the experiment when external forces are neglected and included, respectively. Although, there was not a significant change in the position limit avoidance, the inclusion of external forces improves the limitation at velocity level. For instance, the violation of the velocity limit was reduced for joint 5 from circa 40$\%$ to less than 1$\%$. The 1$\%$ limit violation is probably due to inaccuracies in the dynamic model of the robot. The plots of the external forces show the external torques estimated by the dynamic model and the measured torques by the joint torque sensors of the robot. Despite these perturbations, the joint values stayed between their limits. 

The second experiment evaluates the intuitive reaction of the proposed framework (DCTS) in comparison to the other approaches (OSC and QP-MD). To make a fair comparison, the same force must be applied in the same direction. The task is, as in the first simulation, defined only in 2 coordinates of the rotation. The stiffness matrix $\vecn{K}$ is set to diag(1000) s$^{-1}$ and the damping $\vecn{D}$ to diag(63)s$^{-2}$ to keep the initial orientation during the experiment. To achieve a constant force for all the experiments, a piece with approx. 4.1 Kg of weight is attached to the end effector. The controller does not have knowledge of this piece, i.e., only the weight of the robot structure is considered to compute the gravity compensation. After starting the experiment, the end effector is expected to fall following Newton's law, while holding the orientation.

\begin{figure}
\centering
        \begin{center}
	    \subfigure[]{\includegraphics[scale=0.4]{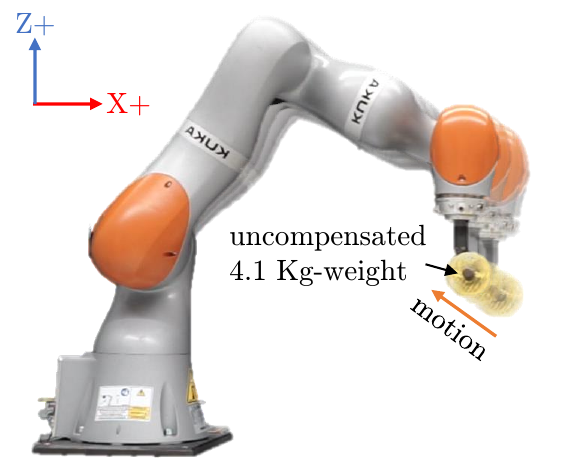}\label{fig:QPVideo}} \subfigure[]{\includegraphics[scale=0.4]{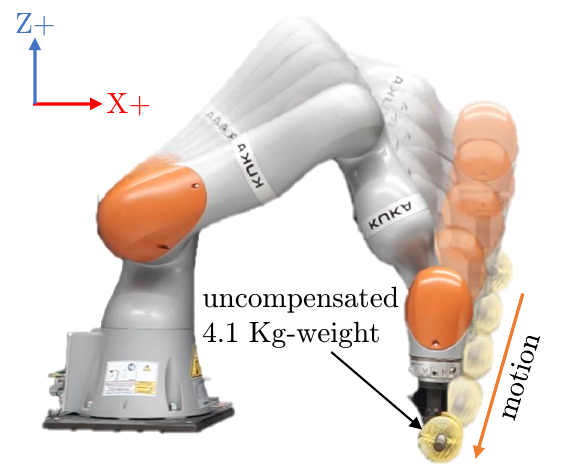}\label{fig:OSCVideo}}
	
        \caption[]{Video snapshots of the performed experiment. (a) Robot's motion with QP-MD. (b) Robot's motion with DCTS or OSC.
        }\label{fig:MotionVideo}
        \end{center}
\end{figure}

\begin{figure}
\centering
        \begin{center}
	    \subfigure[]{\includegraphics[scale=0.52]{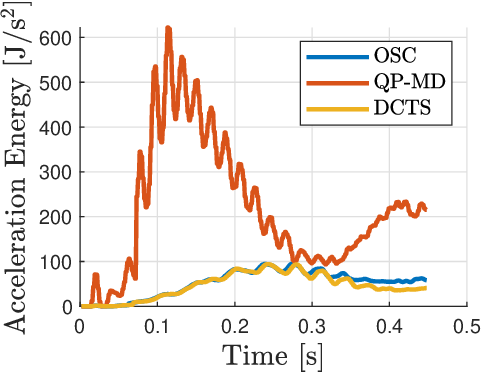}\label{fig:AccEnergyExp}} \subfigure[]{\includegraphics[scale=0.52]{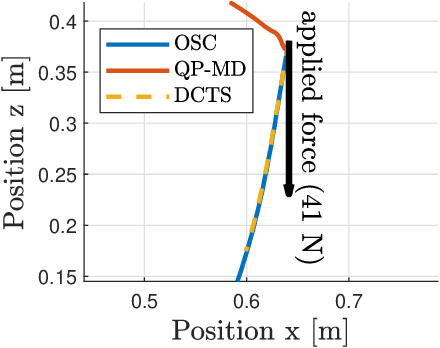}\label{fig:MotionExp1}}
	
        \caption[]{Comparison of Redundancy resolution approaches under physical interaction. a) Acceleration energy. b) Resulting motion in null-space.
        }\label{fig:MotionExp}
        \end{center}
\end{figure}

Fig. \ref{fig:MotionVideo} illustrates the robot's motion after starting the experiment. In the one hand, the intuitive motion of the QP-MD approach is visualized in Fig. \ref{fig:QPVideo}. The end effector starts to move up and backwards instead of falling down. In the other hand, an intuitive motion is shown in Fig. \ref{fig:OSCVideo} achieved by the OSC or DCTS. Note that all approaches kept the desired rotation during the motion. 

The "acceleration energy" plot in Fig. \ref{fig:AccEnergyExp} shows that the QP-MD uses more "acceleration energy" leading to a non- intuitive motion. The plot of the end effector motion in Fig. \ref{fig:MotionExp1} shows that the OSC and the DCTS delivered similar results despite different computation methods.

\section{Conclusion}

This work proposed a framework to perform hierarchical control under unilateral constraints. The framework takes the advantages of  state-of-the-art approaches to solve redundancy as: 
\begin{itemize}
\item the natural solution of approaches based on pseudoinverses and projectors
\item the direct inclusion of unilateral constraints due to the use of QP solvers
\item the avoidance of singularities by the use of QP solvers
\end{itemize} 
Experiments showed that contrary to \cite{SCS}, the DCTS limits the commanded joint generalized forces while computing an optimal solution of the task. In addition, it was demonstrated that the inclusion of the external torques improves the rate of satisfaction of the unilateral constraints. This aspect is very important in industrial robotics. Industrial safety features stop the robot when a position or velocity limit is violated. To avoid an interrupted used of the robot when having physical interaction, the limits must be respected. Experiments showed that a difference of 2$\%$ between the limit in the safety configuration and the limit given to the framework would be enough to allow a continuous operation of the robot while being safe. 

Although the computational effort was not deeply investigated in this paper, the DCTS is expected to be computationally slower than the other solvers. This approach requires the computation of the dynamically consistent projectors in addition to solve the quadratic problem. Furthermore, the constraints of the quadratic problem include augmented matrices of the Jacobian, null-space projectors and desired accelerations. Such high-dimensional matrices increase the computational effort to find the optimal result of the command torques.

\bibliographystyle{ieeetr}  
\bibliography{masterbib}

\end{document}